\pdfoutput=1
\documentclass[11pt]{article}

\usepackage{acl}

\usepackage{times}
\usepackage{latexsym}
\usepackage{algorithm}
\usepackage{algorithmic}
\usepackage{amsmath}
\usepackage{graphicx} 
\usepackage{pifont}
\usepackage{latexsym}
\usepackage{booktabs}
\usepackage{graphicx}
\usepackage{times}
\usepackage{latexsym}
\usepackage{lineno}
\usepackage{caption}
\usepackage{subcaption}
\usepackage{colortbl}
\usepackage{courier}
\usepackage[T1]{fontenc}
\usepackage{amsmath}
\usepackage{amssymb}
\usepackage{multirow}
\usepackage{dsfont}
\usepackage{url}
\usepackage{makecell}
\usepackage{xcolor}
\usepackage{bm}
\usepackage{tikz}
\usepackage{url}
\usepackage{relsize}
\usepackage{hyperref}
\usepackage{enumitem}
\usepackage{makecell}
\usepackage{booktabs}
\usepackage{soul}

\definecolor{lightgreen}{RGB}{197, 224, 180}
\sethlcolor{lightgreen}

\usepackage[T1]{fontenc}
\usepackage[utf8]{inputenc}

\usepackage{microtype}

\title{Unfolding the Headline: Iterative Self-Questioning for \\ News Retrieval and Timeline Summarization}

\makeatletter
\def\@fnsymbol#1{\ensuremath{\ifcase#1\or \dagger\or *\or \ddagger\or
   \mathsection\or \mathparagraph\or \|\or **\or \dagger\dagger
   \or \ddagger\ddagger \else\@ctrerr\fi}}
\makeatother

\author{
Weiqi Wu$^{1,3,4}$\thanks{$~~$This work was done during Weiqi Wu’s internship at Tongyi Lab, Alibaba Group.} ,
Shen Huang$^2$,
Yong Jiang$^2$$^{\ast}$,
Pengjun Xie$^2$,
Fei Huang$^2$,
Hai Zhao$^{1,3,4}$\thanks{$~~$Yong Jiang and Hai Zhao are corresponding authors.} \\
$^1$Department of Computer Science and Engineering, Shanghai Jiao Tong University,\\
$^2$Tongyi Lab, Alibaba Group,\\
$^3$Key Laboratory of Shanghai Education Commission for Intelligent Interaction \\ and Cognitive Engineering, Shanghai Jiao Tong University,\\
$^4$Shanghai Key Laboratory of Trusted Data Circulation and Governance in Web3\\
\texttt{wuwq1022@sjtu.edu.cn,}
\texttt{zhaohai@cs.sjtu.edu.cn}\\
\texttt{\{pangda,yongjiang.jy,chengchen.xpj\}@alibaba-inc.com}
}

\begin{document}
\maketitle
\begin{abstract}
In the fast-changing realm of information, the capacity to construct coherent timelines from extensive event-related content has become increasingly significant and challenging. The complexity arises in aggregating related documents to build a meaningful event graph around a central topic. This paper proposes \textbf{CHRONOS} - \textbf{C}ausal \textbf{H}eadline \textbf{R}etrieval for \textbf{O}pen-domain \textbf{N}ews Timeline Summarizati\textbf{O}n via Iterative \textbf{S}elf-Questioning, which offers a fresh perspective on the integration of Large Language Models (LLMs) to tackle the task of Timeline Summarization (TLS). By iteratively reflecting on how events are linked and posing new questions regarding a specific news topic to gather information online or from an offline knowledge base, LLMs produce and refresh chronological summaries based on documents retrieved in each round. Furthermore, we curate Open-TLS, a novel dataset of timelines on recent news topics authored by professional journalists to evaluate open-domain TLS where information overload makes it impossible to find comprehensive relevant documents from the web. Our experiments indicate that CHRONOS is not only adept at open-domain timeline summarization, but it also rivals the performance of existing state-of-the-art systems designed for closed-domain applications, where a related news corpus is provided for summarization.\footnote{The code and dataset are released at \url{https://github.com/Alibaba-NLP/CHRONOS}.}

\end{abstract}

\section{Introduction}

\begin{figure}[ht]
    \centering
    \includegraphics[width=\linewidth]{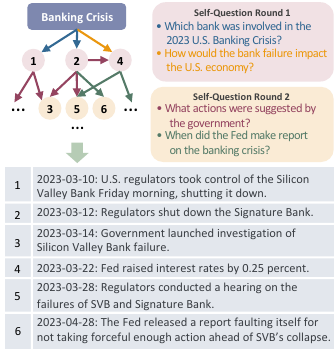}
    \caption{TLS of the news \textit{Banking Crisis}. Edges between event nodes can be established by iterative self-questioning, ultimately building an event graph around the target news for timeline generation. }
    \label{fig:task}
\end{figure}

% Task Introduction + LLM background
The exponential growth of news information in the digital era has made the task of understanding complex event narratives more critical. Timeline Summarization (TLS) \cite{Yan2011EvolutionaryTS, Wang2015OnSA, Chen2019LearningTA, gholipour-ghalandari-ifrim-2020-examining} aims to extract and order the pivotal events from a multitude of textual sources over time, providing a structured view of historical developments. % Despite its significance, the huge amount of data and the difficulty in identifying relevant past news for constructing coherent timelines present considerable challenges to the task.
Despite the complexities inherent in extracting and organizing news events from multiple documents, the advent of Large Language Models (LLMs) \cite{Kojima2022LargeLM, DBLP:journals/corr/abs-2303-08774,DBLP:journals/corr/abs-2309-10305,Bai2023QwenTR} as powerful tools in understanding and generating high-quality text shows their potential in the field of TLS \cite{wang2023webnewstimelinegeneration, hu-etal-2024-moments, Sojitra2024TimelineSI}.

\begin{figure*}[ht]
    \centering
    \includegraphics[width=0.99\linewidth]{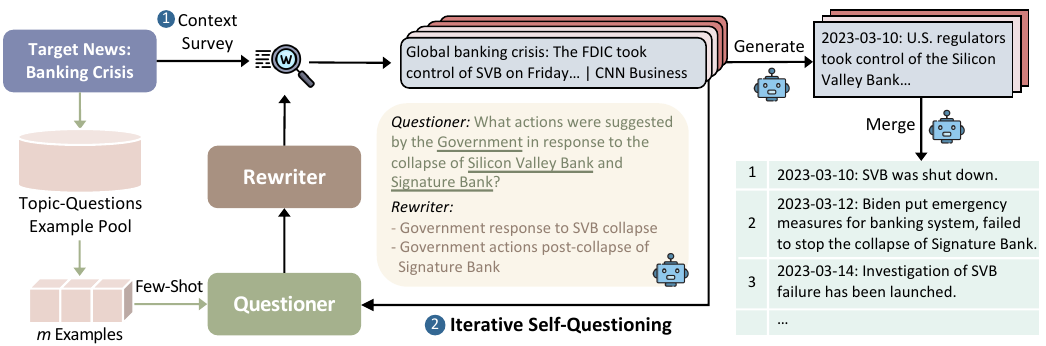}
    \caption{Pipeline of CHRONOS. Giving a target news, it first searches for general context and iteratively poses questions to retrieve more relevant news, while employing a divide-and-conquer strategy to generate the timeline.}
    \label{fig:intro}
\end{figure*}

% causality retrieved by question [motivation] 
The core of synthesizing a timeline is establishing temporal and causal relationships between events \cite{Ansah2019AGI, li-etal-2021-timeline, chen2022follow}. As depicted in Figure \ref{fig:task}, assuming that each news event is represented as a distinct node, our goal is to establish edges between these nodes to present their correlation and ultimately form a heterogeneous graph, starting from the node of topic news. Establishing these edges can be effectively achieved through a search mechanism that retrieves relevant news articles. Thereby, an event node is linked to another if it can retrieve the other event through this search process.

% 讨论不同设置下的chanllenges
Based on the sources of retrievable news, we categorize the TLS task into open-domain and closed-domain settings. Open-domain TLS refers to the process of generating timelines from news directly searched and retrieved from the Internet, while closed-domain TLS involves creating timelines from a predefined set of news articles related to a specific domain. Open-domain TLS faces additional challenges due to the vast and dynamic nature of online information. The information overload makes it difficult to retrieve relevant and comprehensive information from the Internet, introducing noisy data that complicates the task of filtering and assessing the quality of retrieved content. Hence, establishing relationships among events is more challenging in an open domain without access to a global view of relevant news.

% method
To address such challenges, we propose \textbf{CHRONOS}, \textbf{C}ausal \textbf{H}eadline \textbf{R}etrieval for \textbf{O}pen-domain \textbf{N}ews Timeline Summarizati\textbf{O}n via Iterative \textbf{S}elf-Questioning, a new scheme for both settings of TLS based on the Retrieval-Augmented Generation (RAG) framework \cite{Li2022ASO, Zhang2023RetrieveAT, Gao2023RetrievalAugmentedGF, Zhao2024RetrievalAugmentedGF}, as shown in Figure \ref{fig:intro}. By simulating the way humans search for information, which involves learning about the topic by formulating well-defined questions or problems, scanning retrieval results and term suggestions, and further coming up with new subquestions \cite{Bates1989TheDO, ODay1993OrienteeringIA}, we iteratively utilize LLMs to pose 5W1H questions — What, Who, Why, Where, When, How — related to the news topic to gather comprehensive information about related events. We then rewrite the questions to enable a more effective search of it. For each round of retrieved news, we employ an LLM to generate a timeline, which would be merged to produce the ultimate timeline.

% Evaluation
Despite the possibility of evaluating TLS systems in an open-domain setting by not utilizing the corpus provided by current news datasets, these datasets are often limited in size and topic diversity. Therefore, we introduce a more up-to-date and comprehensive news timeline dataset called Open-TLS. It encompasses various topics, including politics, economy, society, sports, and technology, and is sourced from news articles authored by professional journalists.

Our contributions can be summarized as follows:
\begin{itemize}
    \item We propose CHRONOS, a novel retrieval-based approach to TLS by iteratively posing questions about the topic and the retrieved documents to generate chronological summaries.
    \item We construct an up-to-date dataset for open-domain TLS, which surpasses existing public datasets in terms of both size and the duration of timelines.
    \item Experiments demonstrate that our method is effective on open-domain TLS and achieves comparable results with state-of-the-art methods of closed-domain TLS, with significant improvements in efficiency and scalability.
\end{itemize}

\section{Related Works}

\subsection{Timeline Summarization}
Timeline summarization (TLS) synthesizes a chronological narrative of event progression \cite{Allan2001TemporalSO, Chen2019LearningTA, gholipour-ghalandari-ifrim-2020-examining, Yu2021MultiTimeLineS}. While it could be approached as an extension of multi-document summarization \cite{Chieu2004QueryBE, martschat-markert-2018-temporally}, common strategies include focusing pivotal dates \cite{tran-etal-2015-joint, Tran2015TimelineSF, Steen2019AbstractiveTS} or identifying milestone events \cite{li-etal-2021-timeline, chen2022follow}. LLMs have introduced advancements to the field of TLS \cite{wang2023webnewstimelinegeneration, Sojitra2024TimelineSI}. Specifically, \citet{hu-etal-2024-moments} leverage LLMs for the generation and clustering of event summaries.

\subsection{Retrieval-Augmented Generation}
Retrieval-Augmented Generation (RAG) enhances LLMs by incorporating external knowledge during inference, addressing issues such as hallucination and outdated information \cite{liu-etal-2022-makes, shi-etal-2022-nearest, ram-etal-2023-context, JMLR:v24:23-0037, li-etal-2023-unified, agrawal-etal-2023-context}. The retrieval sources of RAG can range from local databases \cite{siriwardhana-etal-2023-improving} to web searches \cite{Nakano2021WebGPTBQ, komeili-etal-2022-internet}. As an application, \citet{shao-etal-2024-assisting} researches a topic via multi-perspective Question-Asking during writing. We focus on the task of TLS and expand it to an open-domain setting, introducing news retrieval using the Internet with new challenges.

\section{Methodology}

We present CHRONOS, a new framework for effective and efficient TLS. It iteratively self-questions about previously retrieved news to gather other related events from various perspectives and combines the timelines it creates from each round of search for a thorough summary.

\subsection{Iterative News Self-Questioning}
The initial step of constructing a timeline for a specific target involves gathering relevant news articles. A straightforward method is to search with the news headline as a keyword to obtain the most general and directly linked information to the target news, where we define the retrieved articles as \textit{News Context}. To obtain more comprehensive information about the target, we ask the LLM to generate questions that cannot be answered based on the news context, and iteratively search for new reference articles according to these questions.

To enhance the quality of self-questioning, we leverage the In-Context Learning (ICL) ability of LLMs by employing a few-shot prompt \cite{FewShotLearners, Dong2022ASO, qian-etal-2024-ape, yao-etal-2024-samples} to instruct the LLM to generate questions about the target news based on the previously retrieved news articles. The few-shot method is known to be highly dependent on the quality of the demonstration examples \cite{liu-etal-2022-makes, yang-etal-2023-demonstration, peng-etal-2024-revisiting}. Therefore, curating effective few-shot examples becomes a critical aspect of our self-questioning method. 

To systematically evaluate the quality of the generated questions in the field of TLS, we introduce the concept of \textit{Chrono-Informativeness} (CI). It is designed to assess the ability of the questions to retrieve relevant documents that align chronologically with a reference timeline produced by a professional journalist. The \textit{Chrono-Informativeness} of a set of questions $Q = (q_1, \dots, q_m)$ for a given news topic is calculated as:
\begin{equation*}
\text{CI}(Q, N) = Date\_F_1(T_{Q, N}, T_{ref})
\end{equation*}

where $T_{Q, N}$ is the timeline generated from the $N$ documents retrieved through the rewritten version of $Q$ (see Sec. \ref{sec:rewrite}), and $T_{ref}$ is the reference timeline. The $Date\_F_1$ score is a widely accepted metric in the field of TLS that compares the dates contained in the generated timeline to those in the reference timeline (detailed in Sec. \ref{sec:metric}).

By generating an extensive set of questions for a given news topic, we can use the greedy algorithm to identify the top $m$ questions that maximize $CI(Q, N)$, selecting the question that provides the greatest improvement in CI during each step. The topic-questions pairs are stored in an example pool. When generating questions for a new target news story, we utilize a BERT-base-uncased model\footnote{\url{https://huggingface.co/google-bert/bert-base-uncased}} to embed the query keyword and apply cosine similarity to retrieve the $s$ most similar topics and associated example pairs from the pool. These 
dynamically retrieved few-shot demonstrations ensure that the demonstrations are contextually relevant and chronologically informative, which enhances the overall quality of the self-questioning process.

\begin{table*}[ht]
\centering
\scalebox{0.85}{
\begin{tabular}{c|c|c|cccccc}
\toprule
\multicolumn{1}{c|}{\multirow{2}{*}{}} & \multirow{2}{*}{\textbf{T17}} & \multirow{2}{*}{\textbf{Crisis}} & \multicolumn{6}{c}{\textbf{OPEN-TLS}}                           \\
\multicolumn{1}{c|}{} &                               &                                  & \textbf{Overall} & Politics & Society & Economy & Sports & Technology  \\ \midrule
\textbf{\# of topics}          & 9                             & 4                                & 50               & 25       & 12      & 5       & 5      & 3    \\
\textbf{\# of timelines}       & 19                            & 22                               & 50               & 25       & 12      & 5       & 5      & 3    \\
\textbf{Avg. \# of articles}   & 508                           & 2310                             & -                & -       & -       & -      & -    & -        \\
\textbf{Avg. \# of pub dates}  & 124                           & 307                              & -                & -       & -       & -      & -    & -        \\
\textbf{Avg. duration (days)} & 212                           & 343                              & 4139             & 4624     & 1719    & 1297    & 8219   & 7694 \\ 
\textbf{Avg. \textit{l}}                & 36                            & 29                               & 23               & 25       & 19      & 22      & 20     & 20 \\
\textbf{Avg. \textit{k}}               & 2.9                           & 1.3                              & 1.8             & 1.8       & 2.1     & 1.7     & 1.8    & 1.6  \\
\bottomrule
\end{tabular}}
\caption{Statistics of closed-domain news TLS datasets and our proposed OPEN-TLS. A timeline contains \textit{l} dates associated with \textit{k} sentences describing the events that happened at each date.}
\label{tab:open-tls}
\end{table*}

\subsection{Question Rewrite}

\label{sec:rewrite}

However, the generated questions are usually quite complex to reach a certain level of depth and breadth, adding difficulty to searching. For instance, regarding the news of the Banking Crisis, the questioner posed a question \textit{What actions were suggested by the government in response to the collapse of Silicon Valley Bank and Signature Bank}, and using this question directly as a query in a search engine yields poor retrieval performance. Hence, we apply a question rewriting mechanism \cite{Ma2023QueryRF} to improve the retrieval precision of our questions, achieved using a few-shot prompt design. Specifically, we employ the LLM to decompose each complex or under-performing query into 2-3 focused queries, such as \textit{Government response to Silicon Valley Bank collapse} and \textit{Government actions post-collapse of Signature Bank}. Such decomposition enhances the specificity and coverage of the retrieved documents, making the subsequent summarization tasks more effective.

\subsection{Timeline Summarization}

To create a coherent timeline containing \textit{l} dates from the news articles retrieved using the questions, we utilize a divide-and-conquer strategy by first generating individual timelines from each round and merging them to produce the final timeline.

\paragraph{Generation} We divide the problem of timeline generation into individual rounds of generation. At the end of each round of self-questioning, % the LLM uses the retrieved documents to identify key events. Since different rounds of retrieval may surface different aspects of the news story, 
the LLM is instructed to extract the significant milestone events with clarified dates and write detailed summarizations of these events, using phrases directly from the news articles when possible to maintain authenticity and accuracy.

\paragraph{Merging} After processing each round individually, the final step is to merge the generated timelines to ensure that only the most significant events are retained. The merging process involves aligning events from different rounds and resolving any conflicts of dates and descriptions. % We first sort all events from the timelines according to their dates in chronological order. Subsequently, 
We instruct the LLM to select the top-\textit{l} milestone events from the original timeline. Dates with more events happening are given precedence as they are likely to be more important since these events are consistently identified across multiple rounds of retrieval.

\section{Open-TLS}

 Evaluating TLS systems commonly involves comparing system-generated timelines to those authored by professional journalists. While several benchmarks have been proposed for closed-domain news TLS along with the provided corpus for each topic, existing public datasets like T17 \cite{T17} and Crisis \cite{Tran2015TimelineSF} remain constrained in terms of size and topical diversity. Furthermore, they often lack the timeliness and flexibility characterized by open-domain timeline generation. To bridge these gaps, we introduce \textit{Open-TLS}, a novel dataset that collects timelines about recent news events, written by professional journalists from reputable news organizations such as the Associated Press\footnote{\url{https://apnews.com}}, Public Broadcasting Service\footnote{\url{https://www.pbs.org}}, and The Guardian\footnote{\url{https://www.theguardian.com}}. 

\begin{table*}[t]
\center
\scalebox{0.85}{
\begin{tabular}{@{}ccccccccc@{}}
\toprule
\multirow{2}{*}{}                    &      & \multicolumn{2}{c}{\textbf{Concat F1}} & \multicolumn{2}{c}{\textbf{Agree F1}} & \multicolumn{2}{c}{\textbf{Align F1}} & \multirow{2}{*}{\textbf{Date F1}} \\
\cmidrule(rl){3-4}
\cmidrule(rl){5-6}
\cmidrule(rl){7-8}
                           &                  & \textbf{R-1}  & \textbf{R-2}  & \textbf{R-1}  & \textbf{R-2} & \textbf{R-1}  & \textbf{R-2} &                          \\ \midrule
                           & \textsc{Direct}  &  0.243           &   0.063          &   0.056          &   0.021         &     0.071        &     0.025       &      0.208                  \\
\textbf{GPT-3.5-Turbo}     & \textsc{Rewrite} &  0.233           &   0.067          &   0.054          &    0.022        &         0.070    &     0.026       &      0.205                  \\
                           & \textsc{CHRONOS} &  0.328           &   0.086          &   0.092          &    0.078        &    0.092        &    0.034        &     0.283           \\ \midrule
                           & \textsc{Direct}  &   0.297          &   0.085          &   0.078          &   0.032         &       0.093      &     0.036       &      0.263                  \\
\textbf{GPT-4o}            & \textsc{Rewrite} &   0.283          &   0.080          &     0.079        &   0.034         &       0.093      &  0.038          &      0.272     \\
                           & \textsc{CHRONOS} &   0.351          &   0.103         & 0.105   &    0.047        & 0.121   &   \textbf{0.051}         &    \textbf{0.343}                    \\ \midrule
                           & \textsc{Direct}  &   0.328          &    0.101         &    0.087         &   0.044         &    0.104         &     0.049       &       0.265                 \\
\textbf{Qwen2.5-72B}         & \textsc{Rewrite} &   0.337        &     0.106        &    0.091         &    0.046        &         0.107    &    0.050        &       0.291                 \\
                           & \textsc{CHRONOS} &  \textbf{0.368}  & \textbf{0.110}   &    \textbf{0.106}         & \textbf{0.049}  &   \textbf{0.125}          & 0.050  &     0.324               \\ \bottomrule
\end{tabular}}
\caption{Experimental results on Open-TLS. We present the outcomes from the optimal self-questioning round.}
\label{tab:res-open}
\end{table*}

 As detailed in Table \ref{tab:open-tls}, Open-TLS comprises 50 timelines across various domains, including politics, economics, society, sports, and technology. The majority of the timelines are published post-2020. Each timeline is accompanied by a publication date and a query keyphrase that facilitates searching. In cases where the news is documented on Wikipedia, the title defined in Wikipedia is used as the query. Otherwise, we manually create a suitable query based on its headline. All timelines are carefully curated to ensure high standards, providing exact dates and accurate narratives.

\section{Experiments}

\subsection{Implementation Details}
We construct experiments on CHRONOS based on three popular LLMs: GPT-3.5-Turbo\footnote{\url{https://platform.openai.com/docs/models/gpt-3-5-turbo}}, GPT-4o\footnote{\url{https://platform.openai.com/docs/models/gpt-4o}}, and Qwen2.5-72B \cite{Bai2023QwenTR}. We report the average results of 3 runs during evaluation.

\paragraph{Example Pool} To build the example pool for the few-shot self-questioning prompt, we utilize GPT-4o to generate 50 questions for topics in the Crisis, T17, and Open-TLS datasets. Each topic is self-questioned based on the directly searched news context. When selecting the most similar demonstrations from the example pool, we exclude the topic-questions pair of the target news.

\paragraph{Search Engine} For open-domain TLS, we use the Bing Web Search API\footnote{\url{https://www.microsoft.com/en-us/bing/apis/bing-web-search-api}} and set the query parameter \textit{freshness} to the publish date of reference timeline to retrieve news articles only before it. We additionally use JINA\footnote{\url{https://jina.ai/reader/}} to read the content of the web pages. In the closed-domain setting, we employ Elasticsearch \cite{Gormley2015ElasticsearchTD}, a well-established text search engine. Each document from the news corpus provided by the dataset is chunked into segments of approximately 500 words for retrieval.

\subsection{Evaluation Metrics}
\label{sec:metric}
We adopt the Tilse framework \cite{martschat-markert-2017-improving, martschat-markert-2018-temporally} to evaluate the generated timeline with reference timelines, which includes the following metrics: 

\paragraph{ROUGE-N} Derived from the original ROUGE-N metrics, these metrics measure the overlap of N-grams in generated and reference timelines: (1) \textit{Concat F1} computes ROUGE by concatenating all date summaries; (2) \textit{Agree F1} computes ROUGE using only summaries of matching dates. (3) \textit{Align F1} initially aligns predicted summaries with reference summaries based on similarity and date proximity, then calculates ROUGE between the aligned summaries, penalizing distant alignments.

\paragraph{Date F1} It is the F1 score of dates in the generated timelines compared with the ground truth.

\subsection{Open-Domain TLS}

\subsubsection{Baselines}

We propose two baselines for Open-Domain TLS. The number of retrieved news by baselines equals the total number of news retrieved by CHRONOS.
\begin{itemize}
    \item \textbf{\textsc{Direct}} Directly search for the target news and output a timeline with the retrieved news.
    \item \textbf{\textsc{Rewrite}} Rewrite the target news to create 2-3 queries, search with these rewritten queries, and output a timeline with the retrieved news.
\end{itemize}

\subsubsection{Results}

The results in Table \ref{tab:res-open} demonstrate a consistent improvement across all metrics when using the CHRONOS approach compared to the baselines for each evaluated model. This indicates that CHRONOS enhances both the quality of event summarization and the alignment of dates with the reference timelines. The higher \textit{Date F1} scores show that CHRONOS is more effective at accurately predicting the correct dates for significant events, with GPT-4o outperforming other models in extracting milestone events. Additionally, the improvements in ROUGE-N metrics suggest that the model excels at producing summaries of news events. Moreover, the general improvement by \textsc{Rewrite} compared with \textsc{DIRECT} shows the advantage of query writing preliminarily.

\subsection{Closed-Domain TLS}

\begin{table}[t]
\centering
\small
\begin{tabular}{@{}ccccc@{}}
\toprule
\textbf{Dataset}                 &      \textbf{Model}        & \textbf{AR-1} & \textbf{AR-2} & \textbf{Date F1} \\ \midrule
\multirow{6}{*}{\textbf{Crisis}} & CLUST         & 0.061         & 0.013         & 0.226            \\
                                 & EGC           & 0.079         & 0.015         & 0.291            \\
                                 & LLM-TLS$^\blacktriangle$       & \textbf{0.112}         & 0.032         & \textbf{0.329}   \\
                                 & LLM-TLS$^\bigstar$       & 0.111         & 0.036         & \underline{0.326}   
                                 \\ \cmidrule(l){2-5} 
                                 & \textsc{Direct}        &  0.094    &   0.031   &    0.182      \\
                                 & \textsc{Rewrite} &  0.093   &  0.040   &   0.215      \\
                                 & \textsc{CHRONOS}     &  \underline{0.108}   &  \textbf{0.045}  &   0.323          \\ \midrule
\multirow{6}{*}{\textbf{T17}}    & CLUST         & 0.082         &  0.020        & 0.407           \\
                                 & EGC           & 0.103	     &  0.024	     & \textbf{0.550}            \\
                                 & LLM-TLS$^\blacktriangle$       & \textbf{0.118}         &  0.036        & 0.528    \\
                                 & LLM-TLS$^\bigstar$       & 0.114         &  \underline{0.040}        & \underline{0.543} 
                                 \\ \cmidrule(l){2-5} 
                                 & \textsc{Direct}        &  0.077      &   0.028      &     0.418      \\
                                 & \textsc{Rewrite} &   0.079     &   0.029    &     0.443      \\
                                 & \textsc{CHRONOS}     &  \underline{0.116}         & \textbf{0.042}         &  0.522           \\ \bottomrule
\end{tabular}
\caption{Comparison of CHRONOS with previous works on closed-domain TLS benchmarks, reporting results of the top model. The best F1 scores are \textbf{bolded}, and the second bests are \underline{underlined}.}
\label{tab:close-res}
\end{table}

\subsubsection{Baselines}

We evaluate CHRONOS on the closed-domain TLS task with several prior event-based approaches:

\begin{itemize}
    \item \textbf{CLUST} \citet{gholipour-ghalandari-ifrim-2020-examining} uses Markov clustering for event aggregation and determines the cluster significance by its date frequency in the news corpus.
    \item \textbf{EGC} \citet{li-etal-2021-timeline} utilizes an event graph modelling method, integrating time-aware optimal transport to compress the whole graph into a salient sub-graph for event selection.
    \item \textbf{LLM-TLS} \citet{hu-etal-2024-moments} leverages LLMs as pseudo-oracles for incremental event clustering to construct timelines from a streaming context. We utilize LLaMA2-13B and Qwen2.5-72B for implementation and denote the resulted systems as LLM-TLS$^\blacktriangle$ and LLM-TLS$^\bigstar$ respectively.
\end{itemize}

\subsubsection{Results}

We select the well-established benchmarks Crisis and T17 for evaluating closed-domain TLS and focus on representative performance metrics including Align F1 (short for AR-1 and AR-2) and Date F1. Table \ref{tab:close-res} presents a comprehensive overview of the performance of CHRONOS alongside previous representative works and two fundamental document retrieval baselines defined in the open-TLS task, i.e., \textsc{DIRECT} and \textsc{REWRITE}. We select the best-performing model to report its performance for presentation. Experiments show that CHRONOS matches and even exceeds the performance of previous models in terms of Alignment-based ROUGE-2 scores on both datasets. For the other lagging indicators, CHRONOS ranks second only to LLM-TLS on the Crisis dataset, as well as its Alignment-based ROUGE-1 score of T17. Regarding Date F1, its performance is less than 0.03 behind the state-of-the-art model, which however suffers from the other two metrics.

\begin{figure*}
    \centering
    \includegraphics[width=0.99\linewidth]{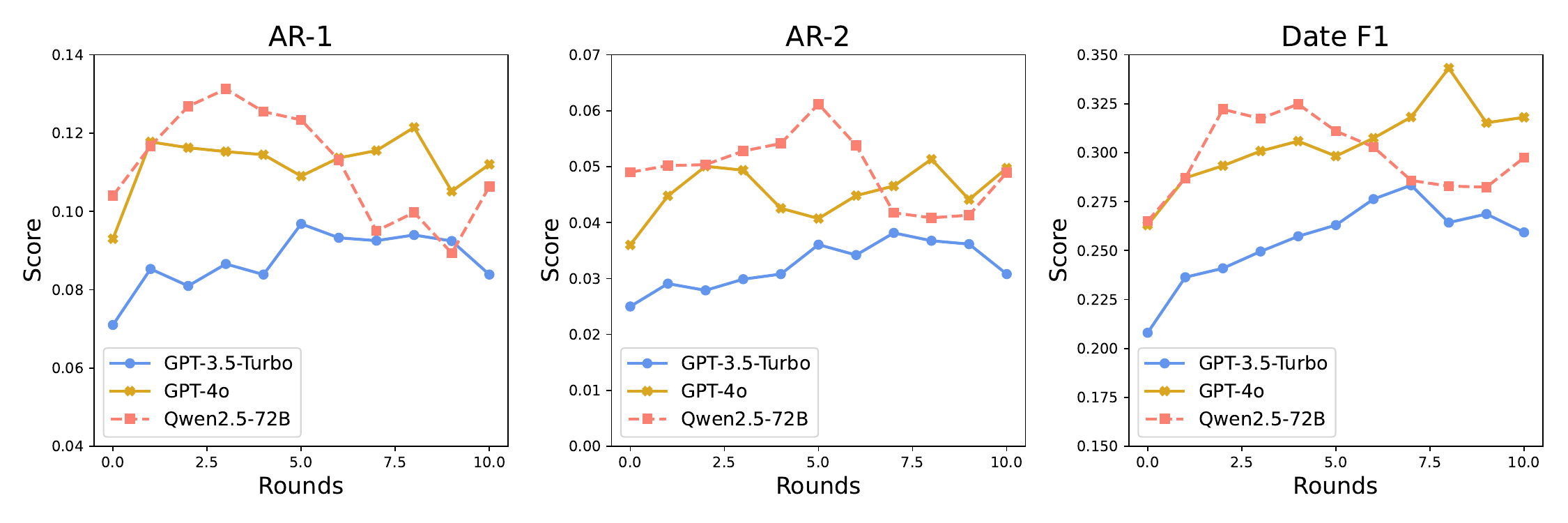}
    \caption{Impact of rounds of Self-questioning on model performance within the Open-TLS dataset.}
    \label{fig:rounds}
\end{figure*}

\subsection{Ablation Study}

\begin{table}[t]
\centering
\small
\begin{tabular}{clccc}
\toprule
\multicolumn{1}{l}{} & \multicolumn{1}{c}{\textbf{Dataset}} & \multicolumn{1}{c}{\textbf{AR-1}} & \multicolumn{1}{c}{\textbf{AR-2}} & \multicolumn{1}{c}{\textbf{Date F1}} \\ \midrule
\multirow{3}{*}{\textsc{CHRONOS}}     & OPEN & \textbf{0.125} & \textbf{0.051} & \textbf{0.343} \\
                             & Crisis   & \textbf{0.108} & \textbf{0.045} & \textbf{0.323} \\
                             & T17      & \textbf{0.116} & \textbf{0.042} & \textbf{0.522} \\ \hline \rowcolor{gray!40} \hline
\multicolumn{5}{l}{\textit{Self-Questioning}}  \\ 
\multirow{3}{*}{\begin{tabular}[c]{@{}c@{}}Random\\ Exemplar\end{tabular}}
                             & OPEN & 0.113 & 0.042 & 0.312 \\
                             & Crisis   & 0.079 & 0.038 & 0.314 \\
                             & T17      & 0.112 & 0.036 & 0.498 \\ \midrule
\multirow{3}{*}{Zero-Shot}   & OPEN & 0.106 &  0.035 &  0.286 \\
                             & Crisis   & 0.059 & 0.023 & 0.306 \\
                             & T17      & 0.102 &  0.037 &  0.471 \\ \hline \rowcolor{gray!40} \hline
\multicolumn{5}{l}{\textit{Question Rewrite}}  \\ 
\multirow{3}{*}{w/o Rewrite} & OPEN & 0.095 & 0.038 & 0.262 \\
                             & Crisis   & 0.078 & 0.047 & 0.286 \\
                             & T17      & 0.072 & 0.026 & 0.446 \\\bottomrule
\end{tabular}
\caption{Ablation study of the topic-questions exemplars and question rewriter. OPEN is short for Open-TLS.}
\label{tab:ab}
\end{table}

\subsubsection{Effects of Question Examples}

CHRONOS selects the top-\textit{s} most similar examples to the target news from the topic-questions example pool to construct few-shot self-questioning prompts. However, when these examples are selected randomly, i.e., \textit{Random Exemplar} in Table \ref{tab:ab}, an evident drop in all metrics is witnessed across the three datasets, demonstrating the effectiveness of strategically selecting examples. This suggests that simply relying on providing examples and neglecting their relevance to the target is suboptimal, as random examples fail to provide contextual guidance for the model. Additionally, using a zero-shot prompt, which bypasses the use of examples entirely, leads to worse performance in most cases. 

\subsubsection{Necessity of Rewriting}

To validate the importance of question rewriting, we compare the performance of our framework with and without this component. As shown in Table \ref{tab:ab}, the removal of the rewriting step leads to a significant decline in the overall performance of TLS, despite a slight improvement ($+0.02$) in the Alignment-based ROUGE-2 score for the Crisis dataset. This minor increase could be due to cases where the original questions closely resemble the phrasing of news articles, which enhances surface-level n-gram overlap. However, the overall decrease in Date F1 and other ROUGE metrics indicates that, without the rewriter, the model encounters difficulties in generating a complete and coherent timeline.

\subsubsection{Rounds of Self-Questioning}

The CHRONOS framework thrives on iterative self-questioning, a process that iteratively expands the news timeline. By increasing the number of questioning rounds, CHRONOS can retrieve a greater volume of news articles, thereby enhancing the comprehensiveness of its news database. However, as depicted in Figure \ref{fig:rounds}, a pattern emerges across all three models on the Open-TLS dataset that their performance initially improves with additional rounds of questioning, but eventually declines. This trend can be attributed to the challenge of merging an excessive number of retrieved news articles into a coherent timeline.

\subsubsection{Number of Retrieved news}

To determine the impact of retrieved news in each round, we experiment with retrieving 20, 30, 40 documents using Qwen2.5-72B on the Open-TLS dataset. Table \ref{tab:N} indicates that increasing the number from 20 to 30 documents significantly improves the results, with marginal improvements when increasing to 40 documents. Intuitively, retrieving more documents provides the model with a richer context. However, due to the potential of introducing noise when integrating less relevant news, the marginal improvements observed when further increasing the number of retrieved news suggest a threshold beyond which the benefits plateau. 

\begin{table*}[h]
\centering
\small
\begin{tabular}{cccccccc}
\toprule
$N$ & \textbf{Concat-R1} & \textbf{Concat-R2} & \textbf{Agree-R1} & \textbf{Agree-R2} & \textbf{Align-R1} & \textbf{Align-R2} & \textbf{Date F1} \\
\midrule
\textbf{20} & 0.321 & 0.082 & 0.078 & 0.041 & 0.098 & 0.042 & 0.287 \\
\textbf{30} & \textbf{0.368} & 0.110 & \textbf{0.106} & \textbf{0.049} & \textbf{0.125} & 0.050 & \textbf{0.324} \\
\textbf{40} & 0.354 & \textbf{0.121} & 0.092 & 0.049 & 0.118 & \textbf{0.051} & 0.321 \\
\bottomrule
\end{tabular}
\caption{Performance on Open-TLS with different numbers of news retrieved in each round.}
\label{tab:N}
\end{table*}

\begin{table}[t]
\centering
\small
\begin{tabular}{@{}ccc@{}}
\toprule
 & \textbf{Crisis}& \textbf{T17}  \\ \midrule
  \textsc{LLM-TLS} &  7 hr 12 min & 2 hr 12 min  \\
 \textsc{CHRONOS} & \textbf{24 min} & \textbf{1 hr 9 min}  \\
\bottomrule
\end{tabular}
\caption{Inference time for LLM-based methods.}
\label{tab:time}
\end{table}

\subsection{Inference Time}

We further compare the running time of the LLM-based methods on the closed-domain datasets, CHRONOS and LLM-TLS. LLM-TLS, which processes each article individually, experiences substantial time delays due to the extensive news corpus of the Crisis dataset. On the other hand, CHRONOS employs a retrieval-based mechanism to focus on highly relevant news articles. Therefore, as shown in Table \ref{tab:time}, CHRONOS spends only 5.6\% of the total time required by LLM-TLS to reach a comparable performance. Even on the T17 dataset with fewer articles per topic, CHRONOS is almost twice as fast while producing similar or improved results. In conclusion, CHRONOS is more practical for real-world applications where efficiency and scalability are critical factors.

\subsection{Discussions}

\subsubsection{Topic Analysis}

 \begin{figure}[t]
    \centering
    \includegraphics[width=\linewidth]{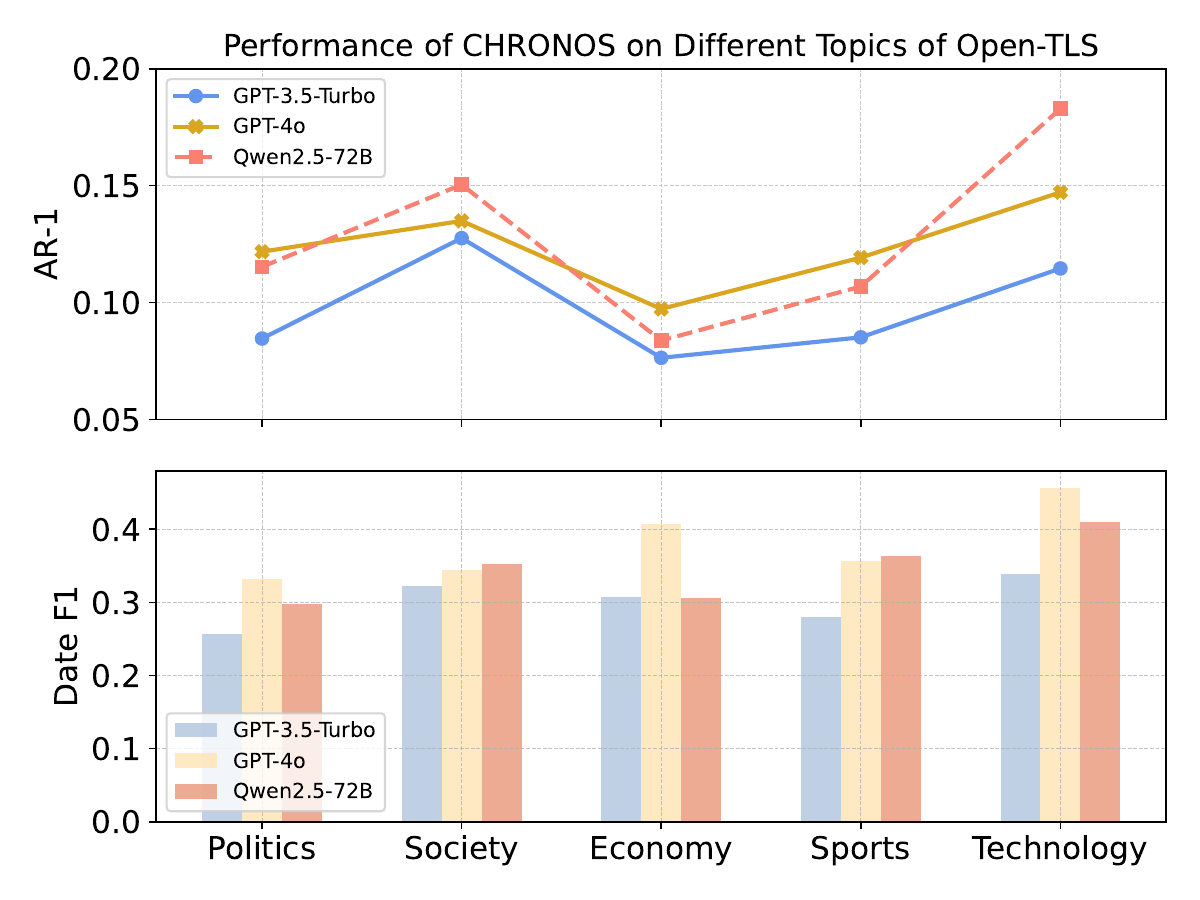}
    \caption{Topic analysis of CHRONOS on Open-TLS.}
    \label{fig:topic}
\end{figure}

We analyze the impact of different topics on the performance of CHRONOS, as shown in Figure \ref{fig:topic}. Upon examining the AR-1 metric, we observe that the Economy and Politics topics tend to challenge the LLMs, likely due to the significant amount of domain knowledge and entities required within these areas. The complexity and specificity of content in these domains make it harder for models to summarize event narratives effectively, resulting in relatively lower scores. Especially for the Economy topic, while the Date F1 scores remain relatively high, indicating that the models are generally successful in extracting dates, the lower ROUGE scores highlight the difficulty of summarizing economic events. Despite the variations in performance across different topics, the three models perform similarly on the Society topic. This convergence in performance could be attributed to the more general and less specialized nature of societal issues, which are easier for the models to handle equally well.

\begin{table*}[t]
\centering
\scalebox{0.9}{
\begin{tabular}{|p{11cm}|p{5cm}|}
\hline
\multicolumn{2}{|c|}{\textbf{Target News: Greatest Apple Announcements (2024.06.30)}} \\ \hline
\multicolumn{1}{|c|}{\textbf{Round 1}}             & \multirow{6}{*}{\parbox{5cm}{\vspace{0.1cm}\textbf{Reference Timeline:}\\\textcolor{ForestGreen}{\textbf{1984-01-24}}: \hl{The Macintosh computer} was unveiled.\\\textcolor{ForestGreen}{\textbf{2001-10-23}}: The \hl{iPod} was unveiled, changing \hl{people}'s view about digital \hl{music} players.\\\textcolor{ForestGreen}{\textbf{2007-01-09}}: The \hl{iPhone} was unveiled, introducing the convenience of touchscreens.\\\textcolor{YellowOrange}{\textbf{2010-01-27}}: The \hl{iPad} was unveiled, creating a middle ground between laptops and smartphones.\\\textcolor{ForestGreen}{\textbf{2014-09-09}}: The \hl{Apple Watch} was unveiled, creating a device that made it possible to wear something akin to a smartphone on your wrist.\\\textcolor{ForestGreen}{\textbf{2016-09-07}}: AirPods was unveiled, helping popularize \hl{wireless headphones} with an Apple chip that provided more reliable and stable connections while making it easy to shift from one gadget to another.\\\textcolor{BrickRed}{\textbf{2024-06-05}}: The Apple Vision Pro was unveiled, which is a mixed-reality headset that seamlessly blends digital content with the physical world.}} \\ \cline{1-1}
\multicolumn{1}{|l|}{\parbox{11cm}{\vspace{0.1cm}\textbf{Self-Question:} \\1. How did Apple transition from early computers into mobile tech?\\2. How has Apple's corporate strategy evolved?\\3. What were the key challenges during the initial years of Apple?\\4. What were the critical factors that led to the innovation of the iPhone?\\5. Who were the individuals instrumental in Apple's major products?\vspace{0.1cm}}}      &                                      \\ \cline{1-1}
\multicolumn{1}{|l|}{\parbox{11cm}{\vspace{0.1cm}\textbf{Generated Timeline:}\\ \textcolor{ForestGreen}{\textbf{1984-01-24}}: Apple introduced the \hl{Macintosh computer}.\\\textcolor{ForestGreen}{\textbf{2007-01-09}}: Apple introduced the first \hl{iPhone}, marking a revolution in the smartphone industry with its multi-touch interface and sleek design.\\\dots\\\textcolor{ForestGreen}{\textbf{2014-09-09}}: Apple announced the iPhone 6 and 6 Plus, as well as the \hl{Apple Watch}, expanding its product lineup beyond phones.\vspace{0.1cm}}} &                                      \\ \cline{1-1}
\multicolumn{1}{|c|}{\textbf{Round 2}}             &                                      \\ \cline{1-1}
\multicolumn{1}{|l|}{\parbox{11cm}{\vspace{0.1cm}\textbf{Self-Question:} \\1. What were the key technological breakthroughs during Apple's development?\\2. How did the launch of the iPhone impact Apple's strategy?\\3. How has the user interface of Apple products evolved?\\4. In what ways has Apple’s design philosophy changed?\\5. How did Apple navigate the transition from Steve Jobs to Tim Cook?\vspace{0.1cm}}}      &                                      \\ \cline{1-1}
\multicolumn{1}{|l|}{\parbox{11cm}{\vspace{0.1cm}\textbf{Generated Timeline:}\\\textcolor{ForestGreen}{\textbf{2001-10-23}}: The \hl{iPod} was launched by Apple, revolutionizing how \hl{people} listened to \hl{music} and laying the groundwork for mobile devices.\\\textcolor{YellowOrange}{\textbf{2010-04-03}}: Apple released the \hl{iPad}, a tablet that combined features from both the iPhone and the Mac.\\\textcolor{ForestGreen}{\textbf{2016-09-07}}: Apple launched the iPhone 7 and 7 Plus, the first models to eliminate the \hl{headphone} jack, emphasizing \hl{wireless} audio.\\\dots\vspace{0.1cm}}} &                                      \\ \hline
\end{tabular}}
\caption{Case study of the timeline generated by CHRONOS. The correctly predicted dates are in \textcolor{ForestGreen}{green}, the missing dates are in \textcolor{BrickRed}{red}, and incorrect dates with correct events are in \textcolor{YellowOrange}{yellow}. Overlapping words between the predicted and reference timeline are also \hl{highlighted}.}
\label{tab:case}
\end{table*}

\subsubsection{Case Study}

Table \ref{tab:case} demonstrates how CHRONOS summarizes a timeline of \textit{Greatest Apple Announcements}, constrained by the news publication date of June 30, 2024. CHRONOS generates two rounds of questions to gradually refine its knowledge of the news from a broad overview to more detailed insights. In Round 1, questions like \textit{How has Apple's corporate strategy evolved?} guide the model to explore Apple's historical milestones and capture key events in it. In Round 2, the questioning shifts toward more specific topics to enrich the timeline with finer details. Comparing the generated timelines from both rounds to the reference timeline, CHRONOS accurately extracts major events with high precision. However, the omission of the Apple Vision Pro announcement and an incorrect date for the iPad unveiling indicate improvement in extracting milestone events with the correct dates.

\section{Conclusion}

In conclusion, this paper presents CHRONOS, a novel framework for TLS that leverages LLMs through an iterative self-questioning and retrieval-based process. Our method addresses the challenge of constructing coherent timelines by systematically retrieving event-related documents, reflecting the causal relationships between events. Experiments demonstrate its effectiveness in both open-domain and closed-domain TLS, as we propose a newly curated Open-TLS dataset for up-to-date open-domain news TLS. Moreover, CHRONOS demonstrates significant improvements in scalability and efficiency, making it a valuable tool for news TLS from vast and unstructured information.

\section*{Limitations}

While our work presents several innovative contributions to the field of TLS, we acknowledge certain limitations that may affect its performance: (1) Our method is heavily dependent on the logical correlation between events for effective retrieval. However, if the causal links between events are not strong enough that they only happened chronologically, the system may struggle to retrieve relevant news articles efficiently. (2) The stability and consistency of our outputs are influenced by the volatility of LLMs and Search Engine Results Pages (SERPs). These fluctuations can lead to variations in the quality and reliability of the summaries generated by CHRONOS in real time.

\section*{Ethics Statement}

A strong commitment to ethical standards and responsible research practices has guided the development and utilization of the Open-TLS dataset. We respect intellectual property rights and the guidelines established by content creators. Hence, we have strictly followed the terms of use set forth by the news organizations and websites from which we sourced the timelines. We have additionally made efforts to construct and present our dataset in a manner that preserves the integrity and accuracy of the original journalistic work. We are dedicated to ensuring that our dataset does not infringe upon the rights or privacy of individuals or organizations. Furthermore, all other datasets and models utilized in this work are publicly accessible and distributed under permissive licenses.

\bibliography{anthology,custom}
\bibliographystyle{acl_natbib}

\appendix

\section{Hyperparameters}

In our experimental configuration, we have set the parameter $m$ to 5, which represents the number of questions that the LLM generates in each round. The parameter $N$ is set to 30, defining the maximum number of retrieved documents in each round. Furthermore, we have designated $s$ as 3, indicating the number of few-shot examples included in the self-questioning prompt.

\section{Prompt Demonstration}

We present the prompts used for three main modules within our system: self-questioning, question rewriting, and timeline generation. 

\subsection{Self-Questioning Prompt}

Table \ref{tab:prompt_questioner} shows the prompt for news self-questioning. With the dynamically selected examples for each target news, the prompt is designed to guide LLMs in formulating a series of questions that expand the scope of the news database for generating the timeline.

\begin{table}[htbp]
    \centering
    \scalebox{0.9}{
    \begin{tabular}{p{\linewidth}}
        \toprule
        \textbf{Instruction for News Self-Questioning}\\
        \midrule
        You are an experienced journalist building a timeline for the target news. You need to propose at least 5 questions related to the Target News that the current news database cannot answer.\\
        These questions should help continue organizing the timeline of news developments or the life history of individuals, focusing on the origins, development processes, and key figures of related events, emphasizing factual news knowledge rather than subjective evaluative content. \\
        These 5 questions must be independent and non-overlapping. The overall potential information volume of all questions should be as large as possible, and the time span covered should also be as extensive as possible. Avoid asking questions similar to those already searched. Directly output your questions in the specified format.\\
        Output format: ["Question\_1", "Question\_2", ...]\\\\
        \{Retrieved Examples\}\\\\
        Current News Database: \{docs\}\\
        Target News: \{news\}\\
        Questions Already Searched: \{questions\}\\
        \bottomrule
    \end{tabular}}
    \caption{Prompt for the questioner.}
    \label{tab:prompt_questioner}
\end{table}

\subsection{Rewrite Prompt}

Table \ref{tab:prompt_rewriter} presents the few-shot prompt used for question rewriting. The examples provided in the prompt demonstrate how to decompose complex questions while preserving their original intent.

\begin{table}[htbp]
    \centering
    \scalebox{0.9}{
    \begin{tabular}{p{\linewidth}}
        \toprule
        \textbf{Instruction for Question Rewriting}\\
        \midrule
        Generate 2-3 rewrite queries of the question as a python list, directly output it as ["..", "..", ..]\\\\
        \# Examples:\\
        Question: When did the initial protests that led to the Egyptian Crisis begin?\\
        Rewrite: ["Egyptian Crisis initial protests", "Time of protests lead to Egyptian Crisis"]\\\\
        Question: When and where did Robert Jasmiden die?\\
        Rewrite: ["Robert Jasmiden's death time", "Robert Jasmiden's death place"]\\\\
        Question: What profession do Nicholas Ray and Elia Kazan have in common?\\
        Rewrite: ["Nicholas Ray profession", "Elia Kazan profession"]\\\\
        Question: \{question\}\\
        Rewrite: \\
        \bottomrule
    \end{tabular}}
    \caption{Prompt for the rewriter.}
    \label{tab:prompt_rewriter}
\end{table}

\subsection{Timeline Generation Prompts}

Table \ref{tab:prompt_generator} and Table \ref{tab:prompt_merger} illustrate the prompts for timeline generation with detailed instructions.

\begin{table}[htbp]
    \centering
    \scalebox{0.9}{
    \begin{tabular}{p{\linewidth}}
        \toprule
        \textbf{Instruction for Timeline Generation}\\
        \midrule
        You are an experienced journalist building a timeline for the target news.\\\\
    Instructions: \\
    Step 1: Read each background news item and extract all significant milestone events related to the target news from your news database, along with their dates. \\
    Step 2: Write a description for each event, including key detail information about the event, using the phrasing from the news database as much as possible. Save all events as a list. The format should be: [\{"start": <date|format as "2023-02-02", cannot be empty, must include specific year, month, and day>, "summary": "<event description|no quotes allowed>"\}, ...] \\\\
    Target News: \{news\}\\
    Current news database: \{docs\}\\
        \bottomrule
    \end{tabular}}
    \caption{Prompt for the timeline generator.}
    \label{tab:prompt_generator}
\end{table}

\begin{table}[htbp]
    \centering
    \scalebox{0.9}{
    \begin{tabular}{p{\linewidth}}
        \toprule
        \textbf{Instruction for Timeline Merging}\\
        \midrule
        You are an experienced journalist building a timeline for the target news.\\
    Merge the existing news summaries and timelines in chronological order. When merging the news summaries, select the top-\{l\} significant news from the original timeline, and strictly follow the chronological order from past to present without changing the original date, using "\textbackslash n" to separate events that occurred on different dates. Directly output your answer in the following format: [\{"start": <date|format as "2023-02-02", cannot be empty, must include specific year, month, and day>, "summary": "<event description|no quotes allowed>"\}, ...]\\

    Target News: \{news\}\\
    Original Timeline: \{timelines\}\\
        \bottomrule
    \end{tabular}}
    \caption{Prompt for merging the timelines from each round.}
    \label{tab:prompt_merger}

\end{table}

\end{document}